# Contrastive and Variational Approaches in Self-Supervised Learning for Complex Data Mining

Yingbin Liang
Northeastern University
Seattle, USA

Lu Dai
University of California, Berkeley
Berkeley, USA

Shuo Shi
Brandeis University
Waltham, USA

Minghao Dai
Columbia University
New York, USA

Junliang Du
Shanghai Jiao Tong University
Shanghai, China

Haige Wang*
University of Miami
Miami, USA

*Abstract-Complex data mining has wide application value in many fields, especially in the feature extraction and classification tasks of unlabeled data. This paper proposes an algorithm based on self-supervised learning and verifies its effectiveness through experiments. The study found that in terms of the selection of optimizer and learning rate, the combination of AdamW optimizer and 0.002 learning rate performed best in all evaluation indicators, indicating that the adaptive optimization method can improve the performance of the model in complex data mining tasks. In addition, the ablation experiment further analyzed the contribution of each module. The results show that contrastive learning, variational modules, and data augmentation strategies play a key role in the generalization ability and robustness of the model. Through the convergence curve analysis of the loss function, the experiment verifies that the method can converge stably during the training process and effectively avoid serious overfitting. Further experimental results show that the model has strong adaptability on different data sets, can effectively extract high-quality features from unlabeled data, and improves classification accuracy. At the same time, under different data distribution conditions, the method can still maintain high detection accuracy, proving its applicability in complex data environments. This study analyzed the role of self-supervised learning methods in complex data mining through systematic experiments and verified its advantages in improving feature extraction quality, optimizing classification performance, and enhancing model stability.*

## I. INTRODUCTION

In today's era of information explosion, data from various domains, including finance systems [1], human-computer interaction [2] and distributed edge computing systems [3-4], is accumulating at an exponential rate. However, these data often exhibit complex, high-dimensional, heterogeneous, and even unstructured characteristics, posing significant challenges to traditional data mining methods. On the one hand, complex data often consist of multiple modalities, such as text, images, speech, and video, each with distinct feature distributions, making information fusion increasingly difficult. On the other hand, a vast amount of data lacks high-quality manual annotations, while supervised learning methods heavily rely on sufficient labeled data. Acquiring such labeled information is not only time-consuming and labor-intensive but may also be infeasible in certain scenarios. Consequently, a crucial research challenge in machine learning is how to efficiently mine information from unlabeled or sparsely labeled data [5].

Self-supervised learning (SSL) has emerged as a significant breakthrough in machine learning in recent years, offering a novel solution for complex data mining [6]. By designing pretraining tasks, SSL enables models to learn meaningful representations from the data itself without relying on manual annotations, thereby reducing the dependence on high-quality labeled data [7]. Compared to traditional supervised learning, SSL can leverage large-scale unlabeled data to discover underlying patterns, enhancing the generalization ability of models, which is particularly advantageous in scenarios where data is scarce or annotation costs are prohibitive [8]. Moreover, SSL can be combined with contrastive learning, generative adversarial networks (GANs), and variational autoencoders (VAEs) to efficiently represent high-dimensional complex data, thereby driving advancements in data mining technologies.

The application of self-supervised learning in complex data mining has shown great promise, especially in the financial field [9-12]. In natural language processing, the successful deployment of large-scale pre-trained models such as BERT and GPT has demonstrated SSL's superiority in learning textual representations. In computer vision and Generative UI Design [13], SSL methods like SimCLR and MoCo leverage contrastive learning strategies to effectively extract image features, improving model performance in recognition and classification tasks [14]. Therefore, exploring complex data mining algorithms based on SSL not only advances AI applications across industries but also provides novel methodologies and technical frameworks for large-scale intelligent data analysis [15-16].

Self-supervised learning (SSL) has progressed significantly in complex data mining but still faces hurdles. Designing effective pretraining tasks remains an open problem, while large-scale, high-dimensional data demand high computational resources, limiting real-world adoption. Integrating multiple data modalities and improving generalization further challenge SSL's development. Nevertheless, SSL holds substantial theoretical and practical value: it offers powerful representation learning on unlabeled data and drives intelligent applications across healthcare, finance, autonomous driving, and beyond [17]. Continued research will reinforce its impact and accelerate industry-wide innovation.

## II. METHOD

This study aims to build a complex data mining algorithm based on self-supervised learning to improve the model's feature learning and generalization capabilities in high-dimensional, heterogeneous, and unlabeled data scenarios. The core idea is to use self-supervised learning methods to learn effective feature representations from unlabeled data and combine contrastive learning and variational inference methods to optimize the model's representation capabilities. The architecture of its feature extractor is shown in Figure 1.

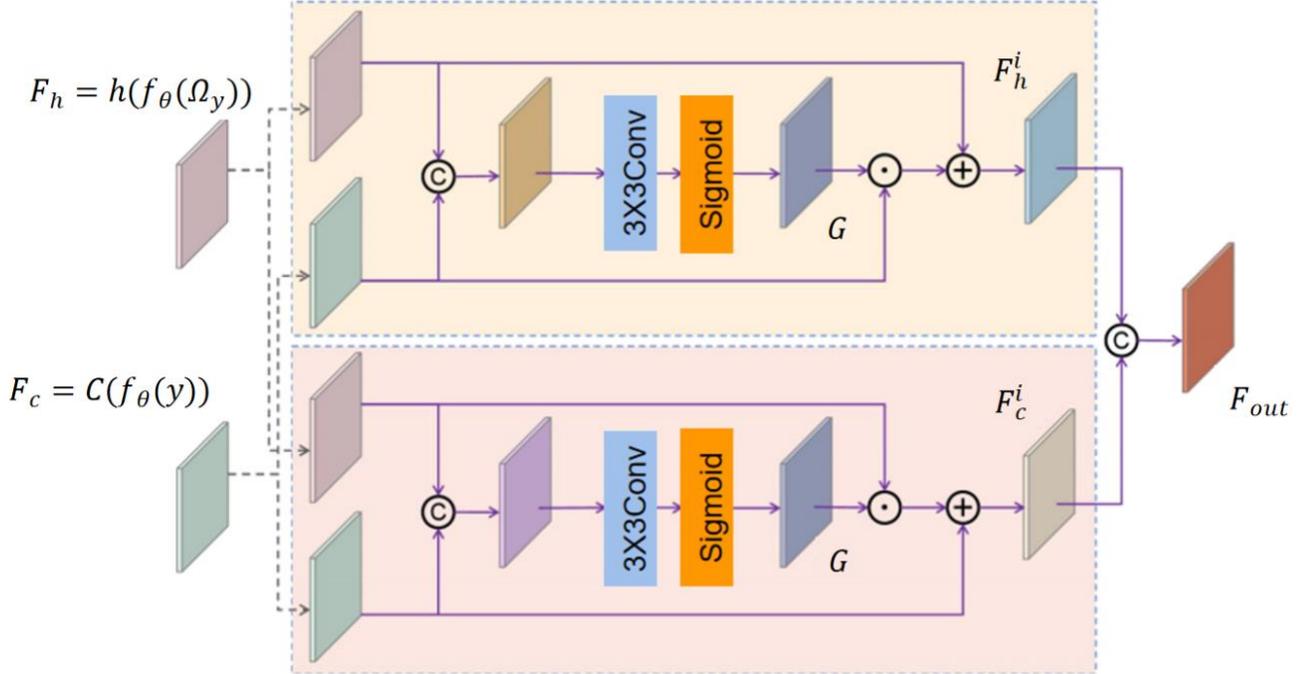

Figure 1. Feature Extractor Architecture

Assume that the dataset $D = \{x_i\}_{i=1}^{N}$ consists of N samples, each sample $x_i$ comes from a high-dimensional complex space $X$. The goal is to learn a mapping function $f : X \rightarrow Z$ under unsupervised conditions to project the input data into a low-dimensional feature space $Z$ while maintaining the semantic consistency and distribution structure of the data.

To achieve this goal, this study constructs a self-supervised learning task inspired by Yan et al [18] employing the structural characteristics inherent in the data to generate pseudo labels. This methodological approach leverages techniques aligned with graph neural networks as described by Qi et al. [19], particularly in scenarios involving complex and imbalanced datasets. Additionally, the incorporation of an attention mechanism, as demonstrated by Li et al. [20], further optimizes the generation of robust and context-aware pseudo labels, enhancing the effectiveness and interpretability of the self-supervised learning framework. By systematically integrating these methodologies, the study aims to effectively exploit underlying structural relationships and patterns in data, ensuring improved accuracy, robustness, and interpretability in downstream analytical tasks. Let $\theta$ be the trainable parameter of the model, and define the self-supervised objective function as maximizing the amount of information of the data in the feature space while maintaining invariance to data transformation. Based on the principles of information theory, this study introduces the concept of mutual information maximization, guided by methodologies presented by Gao et al. [21]. Specifically, we adopt strategies involving transfer and meta-learning to facilitate effective information extraction and representation learning. Additionally, our approach incorporates techniques inspired by fine-tuning Transformer-based models using knowledge graphs, as proposed by Liao et al. [22], to handle complex and nuanced classification tasks. Furthermore, to enhance our model's performance, especially in handling imbalanced datasets, we integrate adaptive weighting strategies drawn from the Markov network classification framework discussed by Wang [23]. Collectively, these methodologies ensure a robust and theoretically grounded approach for optimizing information utilization in the classification process and the feature representation $z_i = f_\theta(x_i)$ fully retains the information of the original data, that is, maximizes the mutual information between the data distribution and the feature representation distribution:

$$L_{SSL} = -\sum_{i=1}^{N} I(x_i; z_i)$$

Since mutual information $I(x_i; z_i)$ is difficult to calculate directly, we adopt a contrastive learning strategy to optimize the model's ability to distinguish in high-dimensional data space by constructing positive and negative sample pairs. Specifically, we use data enhancement methods (such as random cropping, rotation, color jitter, etc.) to generate different perspectives of samples and define a similarity metric function $s(x_i; z_i)$ to measure the closeness of the representations of two samples. The InfoNCE loss function is used as the optimization objective:

$$L_{NCE} = -\sum_{i=1}^{N} \log \frac{\exp(s(z_i, z_i^+)/\tau)}{\sum_{j=1}^{M} \exp(s(z_i, z_j)/\tau)}$$

Among them, $z_i^+$ is the feature representation of the enhanced version of $x_i$, $\tau$ is the temperature parameter, and M is the number of negative samples. The core idea of this loss function is to maximize the similarity of representations of the same data point from different perspectives, while minimizing the similarity with the representations of other data points.

In addition, in order to improve the robustness and data adaptability of feature representation, we further introduce the framework of variational autoencoder (VAE) to model the generative distribution of data through latent variables [24]. Let $q_\phi(z|x)$ be the variational posterior distribution of data and $p_\phi(x|z)$ be the generative distribution of reconstructed data, then the variational lower bound (ELBO) of VAE can be expressed as:

$$L_{VAE} = E_{q\phi(z|x)}[\log p_\theta(x|z)] - D_{KL}(q_\phi(z|x) \| p(z))$$

Among them, $D_{KL}()$ is the Kullback-Leibler divergence, which is used to measure the difference between the variational posterior and the prior distribution $p(z)$. By jointly optimizing $L_{NCE}$ and $L_{NCE}$, the model can learn both discriminative and generative feature representations and enhance the modeling ability of complex data. Finally, we combine the above objective function and define the overall optimization goal as:

$$L = \lambda_1 L_{NCE} + \lambda_2 L_{VAE}$$

Among them, A and B are hyperparameters used to balance the contribution of contrastive learning and variational inference. By optimizing this loss function by gradient descent, we can obtain robust and interpretable feature representations, thereby improving the performance of complex data mining.

## III. EXPERIMENT

### A. Dataeset

This study employs the OpenML CC18 dataset as the experimental benchmark. The CC18 dataset is a high-quality, widely used benchmark dataset for classification tasks, encompassing various complex data types across multiple domains, including finance, healthcare, and text analysis. Curated by the OpenML community, the dataset is characterized by its strong representativeness, moderate data volume, and suitability for evaluating machine learning models. Comprising 72 sub-datasets, each with different dimensionalities, categories, and feature types, CC18 effectively tests the adaptability and generalization capability of self-supervised learning methods in diverse scenarios. Moreover, the dataset presents challenges such as class imbalance, missing values, and uneven category distributions, making it an ideal choice for validating the effectiveness of complex data mining algorithms.

During the experimental process, we applied standardization techniques to the CC18 dataset to ensure consistency in feature scales across different sub-datasets. First, for numerical features, we employed Z-score normalization, transforming them into a distribution with zero mean and unit variance to eliminate scale discrepancies. For categorical features, we utilized one-hot encoding or target encoding to convert discrete values into numerical representations suitable for model training. Additionally, to enhance data robustness, we performed imputation for missing values: numerical features were filled using the mean, while categorical features were filled using the mode. These preprocessing steps ensured data integrity and minimized information loss during model training.

In the experimental design, we split the dataset into 80% training and 20% testing subsets and employed cross-validation to evaluate model performance. To enhance data diversity, we applied data augmentation techniques to the training set, including random noise perturbation and the Synthetic Minority Over-sampling Technique (SMOTE), to balance class distributions and improve model stability when handling imbalanced data. Furthermore, to assess the adaptability of the model across different sub-datasets, we conducted experiments on multiple sub-datasets and computed the average performance metrics, ensuring the generalization capability and applicability of the proposed method on the CC18 dataset.

### B. Experiment result

This paper conducts hyperparameter sensitivity experiments on different hyperparameters, mainly discussing the impact of learning rate and optimizer on model performance. During the experiment, we set different learning rate values and selected a variety of common optimizers for comparison to analyze the impact of different hyperparameter combinations on model convergence speed, final performance and stability. The experimental results are shown in Table 1 and Table 2, which show the importance of hyperparameter selection in the self-supervised learning framework from multiple perspectives and provide a reference for subsequent model optimization.

Table 1.  Hyperparameter sensitivity experiments(Optimizer)

| Optimizer | Acc | F1-Score | Recall | Precision |
|---|---|---|---|---|
| SGD | 0.823 | 0.817 | 0.805 | 0.832 |
| Adam | 0.856 | 0.851 | 0.843 | 0.864 |
| AdamW | 0.864 | 0.860 | 0.852 | 0.871 |

Table 2.  Hyperparameter sensitivity experiments(LR)

| LR | Acc | F1-Score | Recall | Precision |
|---|---|---|---|---|
| 0.005 | 0.842 | 0.838 | 0.829 | 0.851 |
| 0.003 | 0.857 | 0.853 | 0.845 | 0.862 |
| 0.002 | 0.864 | 0.860 | 0.852 | 0.871 |
| 0.001 | 0.835 | 0.832 | 0.824 | 0.843 |

The optimizer experiments demonstrate that both Adam and AdamW significantly outperform SGD across all evaluation metrics. SGD, which uses a basic gradient descent mechanism, achieved an accuracy of 0.823, with lower F1-score, Recall, and Precision. Its sensitivity to gradient fluctuations likely contributed to slower convergence and a higher chance of getting stuck in local minima.

In contrast, Adam's use of momentum and adaptive learning rates led to more stable convergence and better performance. AdamW, which improves upon Adam by properly handling weight decay through decoupled regularization, achieved the best results with an accuracy of 0.864, outperforming all other optimizers. For the learning rate experiments, a value of 0.002 produced the best overall results, reaching 0.864 accuracy and the highest F1-score, Recall, and Precision. This suggests it offers a good balance between convergence speed and training stability. While 0.003 also performed well, 0.005 showed slight instability, and 0.001 underperformed with accuracy dropping to 0.835, indicating that too small a learning rate hampers optimization. In summary, the combination of AdamW and a learning rate of 0.002 yielded the most effective and stable training outcomes. These findings provide practical guidance for optimizer and learning rate selection in similar data mining tasks. The contribution of each model component was further validated through ablation studies, as shown in Table 3.

Table 3. Ablation Experiment Results

| Setting | Acc | F1-Score | Recall | Precision |
|---|---|---|---|---|
| Full Model (Baseline) | 0.864 | 0.860 | 0.852 | 0.871 |
| - w/o Contrastive Loss | 0.842 | 0.837 | 0.828 | 0.850 |
| - w/o Variational Module | 0.849 | 0.845 | 0.838 | 0.857 |
| - w/o Data Augmentation | 0.831 | 0.827 | 0.818 | 0.840 |

Ablation results show that each module contributes to model performance. Removing contrastive loss drops accuracy from 0.864 to 0.842, confirming its role in improving feature discrimination. Without the variational module, accuracy falls to 0.849, showing its benefit in modeling uncertainty. The biggest drop occurs when data augmentation is removed — accuracy drops to 0.831 and F1-score to 0.827, highlighting its importance for robustness and generalization. Overall, contrastive learning and data augmentation are the most critical components. Figure 2 shows the loss curve during training.

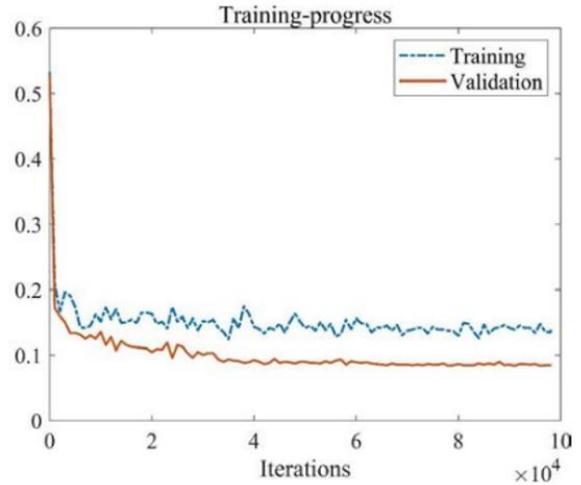

Figure 2. Loss function drop graph

From the loss function decline curve, the model loss value gradually converges during the training process, indicating that the optimization process is effective. At the beginning of training, the loss value is high, indicating that the initial parameters of the model have not yet learned effective features, but with the increase in the number of iterations, the training loss (Training Loss) and the validation loss (Validation Loss) both drop rapidly, indicating that the model is constantly adjusting parameters and gradually fitting the data distribution. After about $1\times10^4$ iterations, the rate of loss decline slows down significantly, and the model enters the stable training stage.

Observing the trends of training loss and validation loss, it can be found that the training loss is always slightly higher than the validation loss, and there is a certain fluctuation. This phenomenon may be due to the randomness of the optimizer or the influence of the data enhancement strategy during the training process, resulting in the loss curve on the training set not being completely smooth. However, overall, the trends of training loss and validation loss are relatively close, and there is no significant divergence, indicating that the model has not been seriously overfitted and can still maintain good generalization ability. Further analysis of the changes in validation loss shows that throughout the training process, validation loss has an overall downward trend and tends to stabilize at a lower level, which indicates that the model has good adaptability to test data. If the validation loss increases significantly in the later stage, it may mean that the model is overfitting and needs to adjust the regularization strategy or early stopping mechanism. In this experiment, validation loss eventually stabilized, indicating that the model has a relatively ideal learning effect on complex data mining tasks.

## IV. CONCLUSION

This study proposed a complex data mining algorithm based on self-supervised learning and verified its effectiveness through a series of experiments. In the sensitivity analysis of optimizers and learning rates, AdamW combined with an appropriate learning rate (0.002) achieved the best performance in all evaluation indicators, indicating the advantages of adaptive optimization methods in complex data mining tasks. At the same time, the ablation experiment results show that contrastive learning, variational modules and data augmentation strategies have significant contributions to the generalization ability and robustness of the model, further verifying the rationality of the proposed method. Through the analysis of the loss function descent curve, we found that the model can converge stably and there is no serious overfitting phenomenon, which proves that this method can effectively extract high-quality feature representations from unlabeled data and improve the accuracy of classification tasks.

Although this study has achieved good experimental results, there are still some limitations. For example, in high-dimensional and multimodal data scenarios, existing methods may not be able to fully mine the correlation information between different modalities, affecting the expressiveness of the model. In addition, although self-supervised learning can be effectively trained on unlabeled data, its migration ability between different tasks still needs to be further explored. Therefore, in future research, it is possible to consider combining multimodal self-supervised learning methods so that the model can establish closer feature associations between data of different modalities, thereby further improving the adaptability and generalization ability of the model. Future research can also optimize large-scale data sets and explore more efficient training strategies, such as combining federated learning, adaptive contrastive learning and other methods to reduce computing costs and improve training efficiency in distributed environments. In addition, with the rapid development of deep learning and generative AI, self-supervised learning can be combined with technologies such as large language models and diffusion models to further improve the performance and application scope of complex data mining. Through these improvements, self-supervised learning can play a greater role in financial risk control, medical diagnosis, intelligent manufacturing, and other fields, providing stronger technical support for intelligent data mining.

## REFERENCES

[1] Y. Cheng, Z. Xu, Y. Chen, Y. Wang, Z. Lin and J. Liu, "A Deep Learning Framework Integrating CNN and BiLSTM for Financial Systemic Risk Analysis and Prediction," arXiv preprint arXiv:2502.06847, 2025.

[2] Q. Sun and S. Duan, "User Intent Prediction and Response in Human-Computer Interaction via BiLSTM," Journal of Computer Science and Software Applications, vol. 5, no. 3, 2025.

[3] J. Zhan, "Elastic Scheduling of Micro-Modules in Edge Computing Based on LSTM Prediction," Journal of Computer Technology and Software, vol. 4, no. 2, 2025.

[4] X. Sun, Y. Yao, X. Wang, P. Li and X. Li, "AI-Driven Health Monitoring of Distributed Computing Architecture: Insights from XGBoost and SHAP," arXiv preprint arXiv:2501.14745, 2024.

[5] X. Shu and Y. Ye, "Knowledge Discovery: Methods from Data Mining and Machine Learning," Social Science Research, vol. 110, p. 102817, 2023.

[6] M. I. Maier, G. Czibula and L. R. Delean, "Using Unsupervised Learning for Mining Behavioural Patterns from Data. A Case Study for the Baccalaureate Exam in Romania," Studies in Informatics and Control, vol. 32, no. 2, pp. 73–84, 2023.

[7] Y. Zhao, C. Zhang, Y. Zhang, et al., "A review of data mining technologies in building energy systems: Load prediction, pattern identification, fault detection and diagnosis," Energy and Built Environment, vol. 1, no. 2, pp. 149–164, 2020.

[8] H. J. Kim and M. K. Kim, "An Unsupervised Data-Mining and Generative-Based Multiple Missing Data Imputation Network for Energy Dataset," IEEE Transactions on Industrial Informatics, 2024.

[9] J. Liu, "Multimodal Data-Driven Factor Models for Stock Market Forecasting," Journal of Computer Technology and Software, vol. 4, no. 2, 2025.

[10] Y. Wang, Z. Xu, Y. Yao, J. Liu and J. Lin, "Leveraging Convolutional Neural Network-Transformer Synergy for Predictive Modeling in Risk-Based Applications," arXiv preprint arXiv:2412.18222, 2024.

[11] X. Du, "Audit Fraud Detection via EfficiencyNet with Separable Convolution and Self-Attention," Transactions on Computational and Scientific Methods, vol. 5, no. 2, 2025.

[12] P. Feng, "Hybrid BiLSTM-Transformer Model for Identifying Fraudulent Transactions in Financial Systems," Journal of Computer Science and Software Applications, vol. 5, no. 3, 2025.

[13] S. Wang, R. Zhang and X. Shi, "Generative UI Design with Diffusion Models: Exploring Automated Interface Creation and Human-Computer Interaction," Transactions on Computational and Scientific Methods, vol. 5, no. 3, 2025.

[14] Y. Huang, Y. Zhang, L. Wang, et al., "Entropystop: Unsupervised deep outlier detection with loss entropy," Proceedings of the 2024 30th ACM SIGKDD Conference on Knowledge Discovery and Data Mining, pp. 1143–1154, 2024.

[15] Y. Deng, "A hybrid network congestion prediction method integrating association rules and LSTM for enhanced spatiotemporal forecasting," Transactions on Computational and Scientific Methods, vol. 5, no. 2, 2025.

[16] M. Chaudhry, I. Shafi, M. Mahnoor, et al., "A systematic literature review on identifying patterns using unsupervised clustering algorithms: A data mining perspective," Symmetry, vol. 15, no. 9, p. 1679, 2023.

[17] M. Zanin, D. Papo, P. A. Sousa, et al., "Combining complex networks and data mining: why and how," Physics Reports, vol. 635, pp. 1–44, 2016.

[18] X. Yan, Y. Jiang, W. Liu, D. Yi and J. Wei, "Transforming Multidimensional Time Series into Interpretable Event Sequences for Advanced Data Mining," Proceedings of the 2024 5th International Conference on Intelligent Computing and Human-Computer Interaction (ICHCI), pp. 126–130, 2024.

[19] Y. Qi, Q. Lu, S. Dou, X. Sun, M. Li and Y. Li, "Graph Neural Network-Driven Hierarchical Mining for Complex Imbalanced Data," arXiv preprint arXiv:2502.03803, 2025.

[20] X. Li, Q. Lu, Y. Li, M. Li and Y. Qi, "Optimized Unet with Attention Mechanism for Multi-Scale Semantic Segmentation," arXiv preprint arXiv:2502.03813, 2025.

[21] J. Gao, S. Lyu, G. Liu, B. Zhu, H. Zheng and X. Liao, "A Hybrid Model for Few-Shot Text Classification Using Transfer and Meta-Learning," arXiv preprint arXiv:2502.09086, 2025.

[22] X. Liao, B. Zhu, J. He, G. Liu, H. Zheng and J. Gao, "A Fine-Tuning Approach for T5 Using Knowledge Graphs to Address Complex Tasks," arXiv preprint arXiv:2502.16484, 2025.

[23] J. Wang, "Markov network classification for imbalanced data with adaptive weighting," Journal of Computer Science and Software Applications, vol. 5, no. 1, pp. 43–52, 2025.

[24] H. S. Jung, H. Lee and J. H. Kim, "Unveiling cryptocurrency conversations: Insights from data mining and unsupervised learning across multiple platforms," IEEE Access, vol. 11, pp. 130573–130583, 2023.